\pdfoutput=1 
\documentclass[10pt,twocolumn,letterpaper]{article}

\usepackage{cvpr}
\usepackage{times}
\usepackage{epsfig}
\usepackage{graphicx}
\usepackage{amsmath}
\usepackage{amssymb}
\usepackage{booktabs}
\usepackage{multirow}
\usepackage{xcolor}
\usepackage{cite}
\definecolor{citecolor}{HTML}{0071bc}

\usepackage[pagebackref=true,breaklinks=true,letterpaper=true,colorlinks,citecolor=citecolor,bookmarks=false]{hyperref}

\cvprfinalcopy 

\ifcvprfinal\fi
\begin{document}

\title{The Semi-Supervised iNaturalist-Aves Challenge at FGVC7 Workshop}

\author{Jong-Chyi Su \quad \quad  Subhransu Maji\\
University of Massachusetts Amherst\\
{\tt\small \{jcsu, smaji\}@cs.umass.edu}
}

\maketitle
\thispagestyle{plain}
\pagestyle{plain}

\begin{abstract}
This document describes the details and the motivation behind a new dataset we collected for the semi-supervised recognition challenge~\cite{semi-aves} at the FGVC7 workshop at CVPR 2020. The dataset contains 1000 species of birds sampled from the iNat-2018 dataset for a total of nearly 150k images. From this collection, we sample a subset of classes and their labels, while adding the images from the remaining classes to the unlabeled set of images. The presence of out-of-domain data (novel classes), high class-imbalance, and fine-grained similarity between classes poses significant challenges for existing semi-supervised recognition techniques in the literature. The dataset is available here: \url{https://github.com/cvl-umass/semi-inat-2020}

\end{abstract}

\section{Introduction}
Recent progress on semi-supervised learning (SSL) for visual recognition (\cite{laine2016temporal,miyato2018virtual,oliver2018realistic,xie2019unsupervised,berthelot2019remixmatch,berthelot2019mixmatch,berthelot2019remixmatch,cascante2020curriculum,xie2020self,zoph2020rethinking,chen2020big,sohn2020fixmatch}) has been hindered by the lack of realistic benchmarks. Datasets such as SVHN~\cite{netzer2011reading}, CIFAR~\cite{krizhevsky2009learning}, STL10~\cite{coates2011analysis}, and ImageNet~\cite{ILSVRC15} that are predominantly used for evaluation suffer from two main shortcomings. 
First, these datasets are \emph{curated} and do not contain significant class imbalance and novel classes that one might see in sparsely labeled datasets observed in real-world applications.
Second, the effect of \emph{transfer learning} is rarely investigated.
Models are either trained from scratch, or the effect of transfer from ImageNet pre-trained models is limited due to the low-resolution or simplicity of the target domain (CIFAR, SVHN, STL10).
In practice, transfer learning is highly effective, especially when transferring to related domains and when learning from a few labels. Thus practical semi-supervised learning should incorporate its benefits.


Motivated by these shortcomings, we collected a new benchmark called \textbf{Semi-Aves} for SSL. The dataset and the challenge was hosted on Kaggle as part of the FGVC7 workshop~\cite{fgvc7} at CVPR 2020~\cite{semi-aves}. Participants were allowed to use ImageNet-1k pretrained models, but not those trained on the iNaturalist dataset~\cite{gvanhorn2018inat}. We next outline the dataset, baselines, and summarize the challenge results. 



\begin{figure}[t]
\centering
\includegraphics[width=\linewidth]{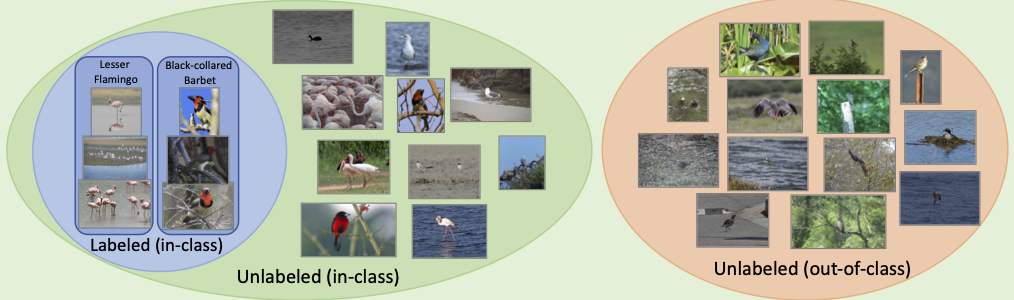}
\caption{The semi-supervised iNaturalist Aves recognition dataset (Semi-Aves) contains sparsely labeled images and unlabeled images from in-class and out-of-class (novel) classes.}
\label{fig:banner}
\end{figure}

\section{The Semi-Aves Recognition Dataset}
We first selected 1000 Aves (birds) species from 1258 Aves taxa included in the iNat-2018 challenge~\cite{inat_2018}. From the 1000 species, we selected 200 species as \emph{in-class} and 800 species as \emph{out-of-class} categories. This allows us to construct unlabeled data with novel classes, as illustrated in Fig.~\ref{fig:banner}.
The images were obtained from iNaturalist website~\cite{iNaturalist} that were uploaded after Februrary 2020, which ensures that there is no overlap with the previous iNat challenge datasets. However, there is a significant class overlap.

For in-class categories, we selected 3,959 images as labeled data ($L_{in}$) and the remaining 26,640 images as in-class unlabeled data ($U_{in}$). We ensure that each class has at least 5 labeled images.
However, due to the natural class imbalance in the original data the ratio of labeled and unlabeled images varies across categories. 
The out-of-class unlabeled data ($U_{out}$) are from the remaining 800 bird species.

The entire collection of in-class and out-of-class unlabeled data could be viewed as unlabeled data in the presence of novel classes. This introduces a domain shift in the labeled and unlabeled set rarely seen in previous benchmarks. Note that we provided the ground-truth labels for in-class or out-of-class for each unlabeled image for the challenge. 

\begin{figure}[h!]
\centering
\includegraphics[width=\linewidth]{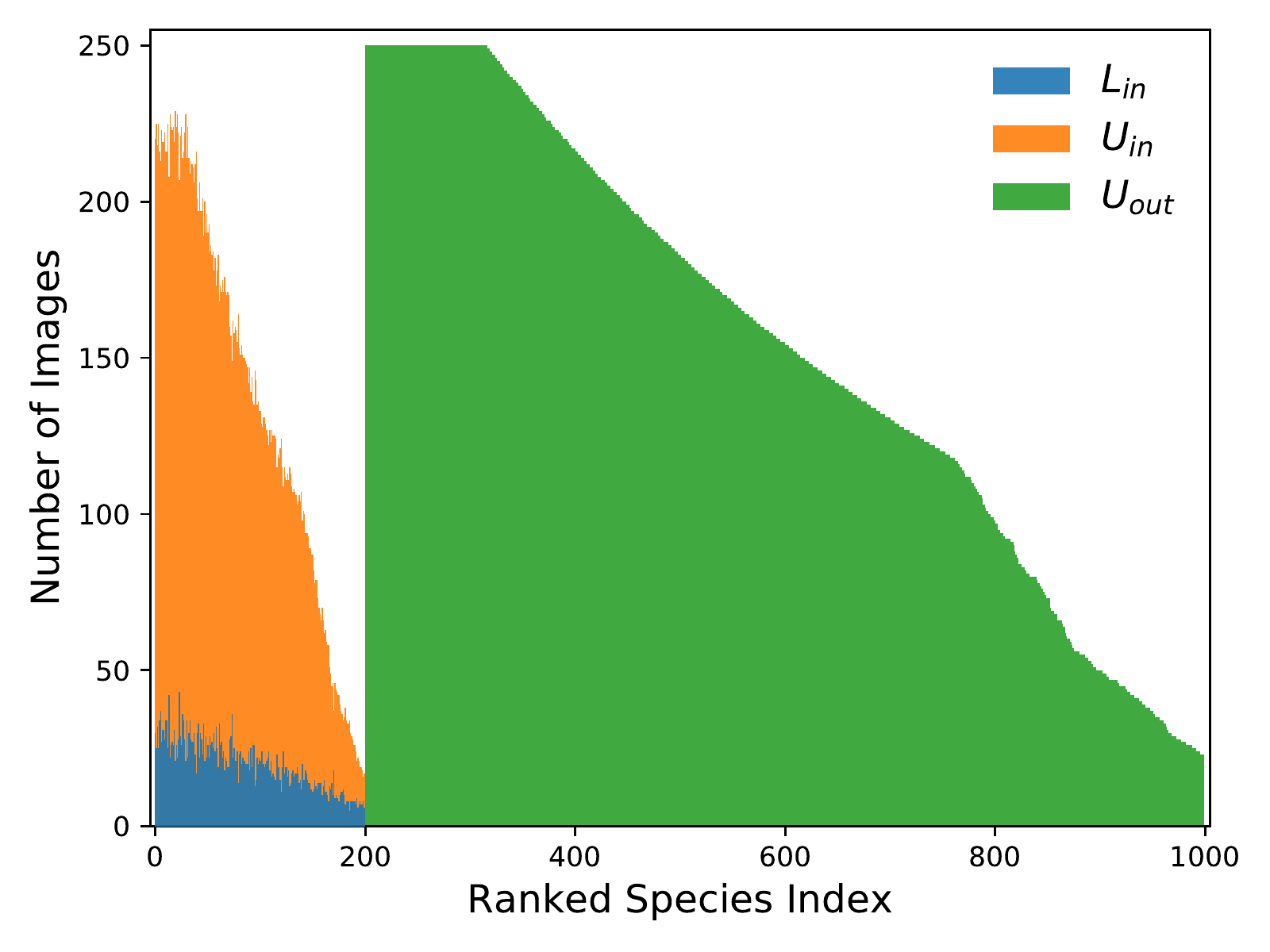}
\caption{Class distribution of Semi-Aves dataset.}
\label{fig:hist}
\end{figure}

Each class has 23 to 250 images and has 122,208 images in total. The validation and test sets have a uniform distribution across categories with 10 and 40 images per class. Half of the test set are used for the public leaderboard and the rest are used to decide the final ranking. The summary of the dataset splits are shown in Tab.~\ref{tab:summary} and the distribution of the images are shown in Fig.~\ref{fig:hist}.
\begin{table}[h!]\label{tab:summary}
  \setlength{\tabcolsep}{7pt}
  \renewcommand{\arraystretch}{1.2}
  \centering
  \begin{tabular}{c c c c c}
    \toprule
Split&	Details&	Classes&	Images\\
    \midrule
    Train&	Labeled&	200&	3,959\\
Train&	Unlabeled, in-class&	200&	26,640\\
Train&	Unlabeled, out-of-class&	800&	122,208\\
Val&	Labeled&	200&	2,000\\
Test&	Public&	200&	4,000\\
Test&	Private&	200&	4,000\\
    \bottomrule
    \end{tabular}
    \caption{Statistics of the Semi-Aves dataset.}
\end{table}

\section{Baselines}
We use ResNet-50~\cite{he2016deep} models trained on various subsets of the data as our baselines, as shown in Tab.~\ref{tab:result}. We use an image resolution of 224$\times$224 --- though higher resolutions improve performance, the trends are similar. The baseline model is obtained by training the model using the cross-entropy loss on the labeled portion of the data only. The model initialized from ImageNet works significantly better than one trained from scratch (14.2\% $\rightarrow$ 43.3\% Top-1 performance). An oracle performance is obtained by training the same model on the entire labeled data. Though the performance from scratch improves to 60.0\%, the transfer learning is still better at 69.0\%.
The gap between the performance suggests that there is a large room for improvement even with a transfer learning baseline.

\begin{table}[h!]\label{tab:result}
  \setlength{\tabcolsep}{9.5pt}
  \renewcommand{\arraystretch}{1.2}
  \centering
  \begin{tabular}{c | c c | c c }
    \toprule
    \multirow{2}{*}{Method} & \multicolumn{2}{c|}{\textbf{from scratch}} & \multicolumn{2}{c}{\textbf{from ImageNet}} \\
     & Top-1 & Top-5 & Top-1 & Top-5 \\
    \midrule
    Baseline & 14.2\%&	32.1\%&	43.3\%&	71.1\%\\
    Oracle & 60.0\%&	82.1\%&	69.0\%&	88.9\%\\
    \bottomrule
    \end{tabular}
    \caption{Performance of ResNet-50 on Semi-Aves.}
\end{table}

\section{Kaggle leaderboard}
We hosted the competition on Kaggle from March to May in 2020. Among the 31 participants, the top team achieves 90.1\% Top1 accuracy on the test set. The winners were asked to present their findings and here we summarize some common trends among the winning methods.
First, most of the teams found pseudo-labeling on in-class unlabeled data to be the most effective. 
State-of-the art methods such as MixMatch~\cite{berthelot2019mixmatch} or FixMatch~\cite{sohn2020fixmatch} were not widely used, or were reported to not provide significant improvement.
Most teams found that there is less benefit from using out-of-class unlabeled data, while one team reported using clustering labels for pre-training is better than only using ImageNet pre-trained model~\cite{cui2020semisupervised}. 
This was surprising to us as the out-of-class data is similar in domain.
Overall, general tricks for image classification such as extensive data augmentation and ensembles were most useful. 

\section{Conclusion}
We have collected a Semi-Aves dataset for evaluating semi-supervised learning in a realistic setting. Comparing to the existing SSL benchmarks, our datasets have distribution mismatch of unlabeled and labeled data, significant class-imbalance, and fine-grained categories. 
A detailed analysis of existing SSL methods and the impact of transfer learning will be described in a forthcoming paper. We are also in the process of organizing the second semi-supervised recognition challenge~\cite{semi-inat} at the FGVC8 workshop~\cite{fgvc8} at CVPR 2021.

\paragraph{Acknowlegements.} This project is supported in part by NSF \#1749833. We also thank the FGVC organization team, Kaggle team for hosting and organizing the workshop, and Oisin Mac Aodha and Grant van Horn for the help collecting the dataset.

{\small
\bibliographystyle{ieee_fullname}
\bibliography{egbib}

\begin{thebibliography}{10}\itemsep=-1pt

\bibitem{iNaturalist}
{\em iNaturalist}.
\newblock \url{https://www.inaturalist.org}.

\bibitem{inat_2018}
{\em iNaturalist 2018 Competition}.
\newblock \url{https://github.com/visipedia/inat_comp/tree/master/2018}.

\bibitem{semi-aves}
{\em Semi-Supervised Fine-Grained Recognition Challenge at FGVC7}.
\newblock
  \url{https://sites.google.com/view/fgvc7/competitions/semisupervised}.

\bibitem{semi-inat}
{\em Semi-Supervised Fine-Grained Recognition Challenge at FGVC8}.
\newblock \url{https://sites.google.com/view/fgvc8/competitions/semi-inat2021}.

\bibitem{fgvc7}
{\em The Seventh Workshop on Fine-Grained Visual Categorization at CVPR}, 2020.
\newblock \url{https://sites.google.com/view/fgvc7}.

\bibitem{fgvc8}
{\em The Eighth Workshop on Fine-Grained Visual Categorization at CVPR}, 2021.
\newblock \url{https://sites.google.com/view/fgvc8}.

\bibitem{berthelot2019remixmatch}
David Berthelot, Nicholas Carlini, Ekin~D Cubuk, Alex Kurakin, Kihyuk Sohn, Han
  Zhang, and Colin Raffel.
\newblock Remixmatch: Semi-supervised learning with distribution alignment and
  augmentation anchoring.
\newblock {\em ICLR}, 2020.

\bibitem{berthelot2019mixmatch}
David Berthelot, Nicholas Carlini, Ian Goodfellow, Nicolas Papernot, Avital
  Oliver, and Colin~A Raffel.
\newblock Mixmatch: A holistic approach to semi-supervised learning.
\newblock In {\em NeurIPS}, 2019.

\bibitem{cascante2020curriculum}
Paola Cascante-Bonilla, Fuwen Tan, Yanjun Qi, and Vicente Ordonez.
\newblock Curriculum labeling: Self-paced pseudo-labeling for semi-supervised
  learning.
\newblock {\em AAAI}, 2021.

\bibitem{chen2020big}
Ting Chen, Simon Kornblith, Kevin Swersky, Mohammad Norouzi, and Geoffrey
  Hinton.
\newblock Big self-supervised models are strong semi-supervised learners.
\newblock {\em NeurIPS}, 2020.

\bibitem{coates2011analysis}
Adam Coates, Andrew Ng, and Honglak Lee.
\newblock An analysis of single-layer networks in unsupervised feature
  learning.
\newblock In {\em Proceedings of the fourteenth international conference on
  artificial intelligence and statistics}, pages 215--223, 2011.

\bibitem{cui2020semisupervised}
Cheng Cui, Zhi Ye, Yangxi Li, Xinjian Li, Min Yang, Kai Wei, Bing Dai, Yanmei
  Zhao, Zhongji Liu, and Rong Pang.
\newblock Semi-supervised recognition under a noisy and fine-grained dataset.
\newblock {\em arXiv preprint arXiv:2006.10702}, 2020.

\bibitem{he2016deep}
Kaiming He, Xiangyu Zhang, Shaoqing Ren, and Jian Sun.
\newblock Deep residual learning for image recognition.
\newblock In {\em CVPR}, 2016.

\bibitem{krizhevsky2009learning}
Alex Krizhevsky, Geoffrey Hinton, et~al.
\newblock Learning multiple layers of features from tiny images.
\newblock 2009.

\bibitem{laine2016temporal}
Samuli Laine and Timo Aila.
\newblock Temporal ensembling for semi-supervised learning.
\newblock {\em ICLR}, 2017.

\bibitem{miyato2018virtual}
Takeru Miyato, Shin-ichi Maeda, Masanori Koyama, and Shin Ishii.
\newblock Virtual adversarial training: a regularization method for supervised
  and semi-supervised learning.
\newblock {\em IEEE transactions on pattern analysis and machine intelligence},
  41(8):1979--1993, 2018.

\bibitem{netzer2011reading}
Yuval Netzer, Tao Wang, Adam Coates, Alessandro Bissacco, Bo Wu, and Andrew~Y
  Ng.
\newblock Reading digits in natural images with unsupervised feature learning.
\newblock {\em NeurIPS Workshop on Deep Learning and Unsupervised Feature
  Learning}, 2011.

\bibitem{oliver2018realistic}
Avital Oliver, Augustus Odena, Colin~A Raffel, Ekin~Dogus Cubuk, and Ian
  Goodfellow.
\newblock Realistic evaluation of deep semi-supervised learning algorithms.
\newblock In {\em NeurIPS}, 2018.

\bibitem{ILSVRC15}
Olga Russakovsky, Jia Deng, Hao Su, Jonathan Krause, Sanjeev Satheesh, Sean Ma,
  Zhiheng Huang, Andrej Karpathy, Aditya Khosla, Michael Bernstein,
  Alexander~C. Berg, and Li Fei-Fei.
\newblock {ImageNet Large Scale Visual Recognition Challenge}.
\newblock {\em International Journal of Computer Vision (IJCV)},
  115(3):211--252, 2015.

\bibitem{sohn2020fixmatch}
Kihyuk Sohn, David Berthelot, Chun-Liang Li, Zizhao Zhang, Nicholas Carlini,
  Ekin~D Cubuk, Alex Kurakin, Han Zhang, and Colin Raffel.
\newblock Fixmatch: Simplifying semi-supervised learning with consistency and
  confidence.
\newblock In {\em NeurIPS}, 2020.

\bibitem{gvanhorn2018inat}
Grant Van~Horn, Oisin Mac~Aodha, Yang Song, Yin Cui, Chen Sun, Alex Shepard,
  Hartwig Adam, Pietro Perona, and Serge Belongie.
\newblock The {iNaturalist} species classification and detection dataset.
\newblock In {\em CVPR}, 2018.

\bibitem{xie2019unsupervised}
Qizhe Xie, Zihang Dai, Eduard Hovy, Minh-Thang Luong, and Quoc~V Le.
\newblock Unsupervised data augmentation for consistency training.
\newblock {\em NeurIPS}, 2020.

\bibitem{xie2020self}
Qizhe Xie, Minh-Thang Luong, Eduard Hovy, and Quoc~V. Le.
\newblock Self-training with noisy student improves imagenet classification.
\newblock In {\em CVPR}, 2020.

\bibitem{zoph2020rethinking}
Barret Zoph, Golnaz Ghiasi, Tsung-Yi Lin, Yin Cui, Hanxiao Liu, Ekin~D Cubuk,
  and Quoc~V Le.
\newblock Rethinking pre-training and self-training.
\newblock {\em NeurIPS}, 2020.

\end{thebibliography}
}

\end{document}